\documentclass{article}

\usepackage[square,numbers]{natbib}

\usepackage[preprint]{arxiv_2021}

\usepackage[utf8]{inputenc} 
\usepackage[T1]{fontenc}    
\usepackage{hyperref}       
\usepackage{url}            
\usepackage{booktabs}       
\usepackage{amsfonts}       
\usepackage{nicefrac}       
\usepackage{microtype}      
\usepackage{xcolor}         

\usepackage{graphicx}
\usepackage{amsmath}
\usepackage{amssymb}
\usepackage{multirow}
\usepackage{comment}
\newcommand{\cmark}{\checkmark}%
\newcommand{\xmark}{\times}%

\title{Referring Transformer: A One-step Approach to Multi-task Visual Grounding}

\author{   
    \makebox[2in][c]{Muchen Li$^{1,2}$} \makebox[2in][c]{Leonid Sigal$^{1,2,3,4}$}  \\
    \makebox[2in][c]{\texttt{muchenli@cs.ubc.ca}}
    \makebox[2in][c]{\texttt{lsigal@cs.ubc.ca}} \vspace{0.05in} \\ 
    $^1$Department of Computer Science, University of British Columbia \\
    $^2$Vector Institute for AI  ~~~~~~
    $^3$CIFAR AI Chair ~~~~~~ $^4$NSERC CRC Chair
}

\begin{document}

\maketitle

\begin{abstract}
\quad As an important step towards visual reasoning, visual grounding (e.g., phrase localization, referring expression comprehension / segmentation) has been widely explored.
Previous approaches to referring expression comprehension (REC) or segmentation (RES) either suffer from limited performance, due to a two-stage setup, or require the designing of complex task-specific one-stage architectures. 
In this paper, we propose a simple one-stage multi-task framework for visual grounding tasks.
Specifically, we leverage a transformer architecture, where two modalities are fused in a visual-lingual encoder. In the decoder, the model learns to generate contextualized lingual queries which are then decoded and used to directly regress the bounding box and produce a segmentation mask for the corresponding referred regions.
With this simple but highly contextualized model, we outperform state-of-the-art methods by a large margin on both REC and RES tasks. 
We also show that a simple pre-training schedule (on an external dataset) further improves the performance. 
Extensive experiments and ablations illustrate that our model benefits greatly from contextualized information and multi-task training.
\end{abstract}

\section{Introduction}

Multi-modal {\em grounding}\footnote{Grounding and referring  expression comprehension have been used interchangeably in the literature. While the two terms are indeed trying to characterize the same task of associating a lingual phrase with an image region, there is a subtle difference in that referring expressions tend to be unique and hence need to be grounded to a single region, {\em e.g.}, ``{\em person in a red coat next to a bus}". The grounding task, as originally defined in \cite{plummer2015flickr30k}, is more general where a lingual phrase may be ambiguous and therefore grounded to multiple regions, {\em e.g.}, ``{\em a person}".} tasks ({\em e.g.}, phrase localization \cite{bajaj2019g3raphground,chen2017query,dogan2019neural,plummer2015flickr30k,wang2018learning}, referring expression comprehension
\cite{hong2019learning,hu2017modeling,kazemzadeh2014referitgame,liao2020real,liu2019learning, liu2019improving,mao2016generation,yang2019fast, yang2020improving,yu2016modeling,yu2018mattnet} and segmentation 
\cite{chen2019referring, hu2016segmentation, hu2020bi, huang2020referring, liu2019learning, margffoy2018dynamic, ye2019cross, yu2018mattnet}) aim to generalize traditional object detection and segmentation to localization of regions (rectangular or at a pixel level) in images that correspond to free-form linguistic expressions.
These tasks have emerged as core problems in vision and ML due to the breadth of applications that can make use of such techniques, spanning image captioning, visual question answering, visual reasoning and others.

The majority of multi-modal grounding architectures, to date, take the form of two-stage approaches, inspired by Faster RCNN \cite{faster_rcnn} and others, which first generate a set of image region proposals and then associate/ground one, or more, of these regions to a phrase by considering how well the content matches the query phrase. Context among the regions and multiple query phrases, which often come parsed from a single sentence, has also been considered in various ways ({\em e.g.}, using LSTM stacks \cite{dogan2019neural}, graph neural networks \cite{bajaj2019g3raphground} and others). More recent variants leverage pre-trained multi-modal Transformers ({\em e.g.}, ViLBERT \cite{lu2019vilbert,lu202012}) to fine-tune to the grounding tasks. Such models have an added benefit of being able to learn sophisticated cross-modal feature representations from external large-scale data, which further improve the performance. However, a significant limitation of all such two-stage methods is their inability to condition the proposal mechanism on the query phrase itself, which inherently limits the upper bound of performance (see Table 3 in \cite{yang2019fast}).

To address these limitations, more recently, a number of one-stage approaches have been introduced \cite{jing2021locate,luo2020multi,yang2019fast,yang2020improving}. Most of these take inspiration from Yolo \cite{redmon2018yolov3} and the variants, and rely on more integrated visual-linguistic fusion and a dense anchoring mechanism to directly predict the grounding regions. While this alleviates the need for a proposal stage, it instead requires somewhat ad hoc anchor definitions, often obtained by clustering of labeled regions, and also limits ability to contextualize grounding decisions as each query phrase is effectively processed independently. Finally, little attention in the literature has been given to leveraging relationship among the REC and RES tasks.

In this work we propose an end-to-end one-stage architecture, inspired by the recent DETR \cite{carion2020end} detection framework, which is capable of simultaneous language grounding at both a bounding-box and segmentation level, without requiring dense anchor definitions. This model also enables contextualized reasoning by taking into account the entire image, all referring query phrases of interest and (optionally) lingual context ({\em e.g.}, a sentence from which referring phrases are parsed).
Specifically, we leverage a transfomer architecture, with a visual-lingual encoder, to encode image and lingual context, and a two-headed (detection and segmentation) custom contextualized tranformer decoder. The contextualized decoder takes as input learned contextualized phrase queries and decodes them directly to bounding boxes and segmentation masks. Implicit 1-to-1 correspondence between input referring phrases and resulting outputs also enables a more direct formulation of the loss without requiring Hungarian matching. 
With this simple model we outperform state-of-the-art methods by a large margin on both REC and RES tasks. 
We also show that a simple pre-training schedule (on an external dataset) further improves the performance. 
Extensive experiments and ablations illustrate that our model benefit greatly from the contextualized information and the multi-task training.

{\bf Contributions.}
Our contributions are:
(1) We propose a simple and general one-stage transformer-based architecture for referring expression comprehension and segmentation. The core of this model is the novel transformer decoder that leverages contextualized phrase queries and is able to directly decode those, subject to contextualized image embeddings, into corresponding image regions and segments; 
(2) Our approach is unique in enabling simultaneous REC and RES using a single trained model (the only other method capable of this is \cite{luo2020multi}); showing that such multi-task learning leads to improvements on both tasks; 
(3) As with other transformer-based architectures, we show that pre-training can further improve the performance and both vanila and pre-trained models outperform state-of-the-art on both tasks by significant margins (up to 8.5\% on refcoco dataset for REC and 19.4\% for RES). We also thoroughly validate our design in detailed ablations. 

\vspace{-0.07in}
\section{Related works}
\vspace{-0.07in}
\label{sec:related_works}
{\bf Referring Expression Comprehension (REC).}\quad
REC focuses on producing an image bounding box tightly encompassing a language query.
Previous two-staged works \cite{hong2019learning,hu2017modeling,liu2019learning, liu2019improving,yu2018mattnet} reformulate this as a ranking task with a set of candidate regions predicted from a pre-trained proposal mechanism. 
Despite achieving great success, the performance of two-staged methods is capped by the speed and accuracy of region proposals
in the first stage. 
More recently, one-stage approaches \cite{liao2020real,yang2019fast, yang2020improving} have been used to alleviate the aforementioned limitations. Yang {\em et al.}~\cite{yang2019fast,yang2020improving} proposed to fuse query information with visual features and pick the bounding box with maximum activation scores from YOLOv3~\cite{redmon2018yolov3}. Liao {\em et al.}~\cite{liao2020real} utilizes CenterNet~\cite{duan2019centernet} to perform correlation filtering for region center localization. 
However, such methods either require manually tuned anchor boxes or suffer from semantic loss due to modality misalignment. 
In contrast, our model
learns to better align modalities using a cross-modal transformer and directly decode bounding boxes for each 
query.

{\bf Referring Expression Segmentation (RES).}\quad
Similar to REC, RES, proposed in~\cite{hu2016segmentation}, aims to predict segmentation masks to better describe the shape of the referred region. A typical solution for referring expression segmentation is to fuse multi-modal information with a segmentation network ({\em e.g.}, \cite{mask_rcnn,FCN}) and train it to output the segmented masks~\cite{hu2016segmentation, liu2019learning, margffoy2018dynamic, ye2019cross, yu2018mattnet}. More recent approaches focus on designing module to enable better multi-modal interactions, {\em e.g.}, progressive multi-scale fusion used in~\cite{huang2020referring} and cross-modal attention block used in~\cite{hu2020bi}.
Since localization information matters in predicting instance segmentations (as noted in Mask RCNN \cite{mask_rcnn}),
very recent work~\cite{jing2021locate} aims to explicitly localize object before doing segmentation. 
Despite the relatively high performance being achieved in RES, existing approaches still struggle to determine the correct referent region and tend to output noisy segmentation results with an irregular shape, while our model is able to produce segmentations with fine-grained shapes even on challenging scenarios with occlusions or shadows.

{\bf Multi-task Learning for REC and RES.}\quad
Multi-task learning is widely applied in object detection and segmentation~\cite{carion2020end,mask_rcnn}, often, by leveraging shared backbone and task-specific heads.
Building on this idea, Luo {\em et al.} ~\cite{luo2020multi} proposed a multi-task collaborative network (MCN) to jointly address REC and RES. 
They introduce consistency energy maximization loss that constrains the feature activation map in REC and RES to be similar. 
While our model is also set up to learn REC and RES tasks jointly, we argue that an explicit constraint tends to downplay the quality of the final predicted mask since the feature map from the REC branch can blur out fine-grained region shape information needed by the RES branch 
(see Figure~\ref{fig:quali_result_refcoco}).
Hence, we use an implicit constraint where tasks head of REC and RES are trained to output corresponding bounding box and mask from the same joint multi-modal representation. 
We illustrate that our model can benefit from multi-task supervision, leading to more accurate results as compared to single-task variants.


{\bf Pretrained Multi-modal Transformers.}\quad
Transformer-based pretrained models \cite{chen2020uniter, gan2020large,lu2019vilbert, tan2019lxmert, yu2020ernie} have recently showed strong potential in multi-modal understanding. 
LXMERT \cite{tan2019lxmert} and ViLBERT \cite{lu2019vilbert} use two stream transformers with cross-attention transformer layers on top for multimodal fusion. More recent works, \cite{chen2020uniter,gan2020large} advocate a single-stream design to fuse two modalities earlier.
The success of the aforementioned models can largely be attributed to the cross-modal representations obtained by multi-task pretraining on a large amount of aligned image-text pairs. 
Despite state-of-the-art performance of such models on the downstream REC task, these models, fundamentally, are still a form of a 
two-stage pipeline where image features are extracted using pretrained detectors or proposal mechanisms.
We focus on a one-stage architecture variant that allows visual and lingual features to be aligned at the early stages.
Although the focus of our work is not to design a better pretraining scheme, we show that our model can outperform the existing state-of-the-art with proper pretraining. 

{\bf Transformer-based Detectors.}\quad
More recently, DETR \cite{carion2020end} and its variants \cite{gao2021fast, zhu2020deformable}, were proposed to enable end-to-end object detection. DETR reformulates detection as a set prediction tasks and uses transformers to decode learnable queries to bounding boxes.
Despite state-of-art performance, DETR is disadvantaged by its optimization difficulty and, usually, extra-long training time. 
While adopting a similar pipeline, our model focus on aligning different modalities to generate contextualized expression-specific referring queries. 
We also design our model to get rid of Hungarian matching loss by leveraging one-to-one correspondences between predicted bounding boxes and referring expressions, which leads to faster convergence for our model.

{\bf Contemporaneous Works.}
Concurrent and independent to us, very recently, there are some works that use transformers for visual referring tasks~\cite{deng2021transvg, du2021visual, kamath2021mdetr}. {\em All three of these are non-refereed/unpublished at the time of this manuscript upload to ArXiv.}
Importantly, unlike \cite{deng2021transvg, du2021visual, kamath2021mdetr} our approach formulated in multi-task setting and solves both REC and RES tasks simultaneously; while the aforementioned works \cite{deng2021transvg, du2021visual} focus on REC specifically. 
In addition, our model is faster and is capable of grounding multiple contextualized phrases, while \cite{deng2021transvg,du2021visual} follow previous one-stage approaches and are only able to infer a single expression at a time; leading, in our case, to more accurate results. Finally, \cite{kamath2021mdetr} builds on DETR \cite{carion2020end} 
while our approach deviates from DETR in a number of ways, including in terms of the loss, and focuses more on the end-to-end joint multi-task learning.


\vspace{-0.07in}
\section{Approach}
\vspace{-0.07in}
\label{sec:approch}
%
%

%
%
Given an image $\mathcal{I}$ and a set of query phrases / referring expressions $\mathcal{Q}_p = \{ \mathbf{p}_i \}_{i = 1, ..., M}$, that we assume to come from an (optional) contextual text source\footnote{For example, a sentence from which noun phrases/referring expressions $\mathbf{p}_i$ were parsed. Where contextual text source is unavailable and only one phrase exists, we simply let $\mathcal{Q} = \mathcal{Q}_p$.} $\mathcal{Q}$, 
our goal is to predict a set of bounding boxes $\mathcal{B} = \{ \mathbf{b}_i \}_{i = 1, ..., M}$ and corresponding segmentation masks $\mathcal{S} = \{ \mathbf{s}_i \}_{i = 1, ..., M}$, one for each query phrase $i$ that localizes that phrase in the image.
Note, $M$ is the number of phrases / referring expressions for a given image $\mathcal{I}$ and is typically between $1$ and $16$ for the Flickr30k \cite{plummer2015flickr30k} dataset.


\begin{figure}[!t]
  \centering
  \includegraphics[width=0.98\textwidth]{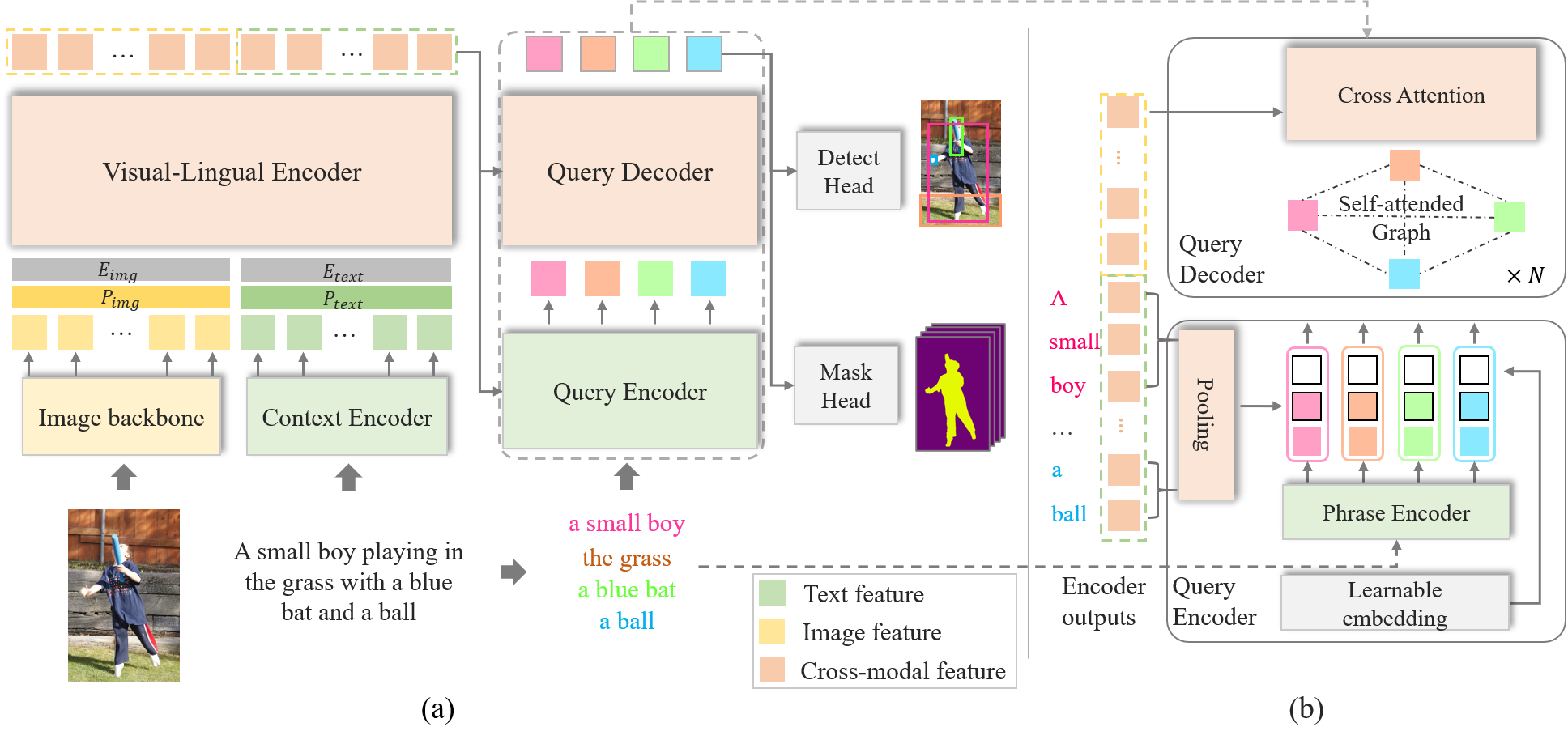}
  \vspace{-0.15in}
  \caption{{\bf Referring Transformer.} An overview of the proposed architecture is shown in (a). For an image and (con)text input, a visual-lingual encoder is used to 
  refine 
  image features, extracted from a convolutional backbone, and lingual features, extracted by a BERT. A query encoder and decoder produce features for REC and RES heads, given multi-modal features and referring expressions. The detailed structure of the query encoder and decoder is shown in (b). Colored squares denote embeddings for corresponding referring queries.}
  \label{fig:net_arch} 
\vspace{-0.1in}
\end{figure}

As shown in Figure~\ref{fig:net_arch}, our referring transformer is composed of four stages.
Given an image-(con)text pair, $<\mathcal{I}, \mathcal{Q}>$, a {\em cross-modal encoder} generates joint image-text embeddings for each visual and textual token -- feature columns and word embeddings respectively. 
Phrases / referring expressions $\mathcal{Q}_p$ and the corresponding pooled joint embeddings (from cross-modal encoder) are then fed into a {\em query generator} which produces phrase-specific queries.
%
The {\em decoder} jointly reasons across all these queries and decodes joint image-(con)text embeddings.
%
The decoded feature is then sent to the {\em detection} and {\em segmentation head} to produce a set of detection boxes $\mathcal{B}$ and segmentation masks $\mathcal{S}$.

%
The result is a one-staged end-to-end model that solves the REC and RES tasks at the same time. We will now introduce constituent architectural components for the four stages briefly described above. 


\subsection{Feature Extraction}
\label{sec:feature_extraction}
{\bf Visual \& Text Backbone.}\quad
Starting from an initial image $\mathcal{I} \in \mathbb{R}^{3 \times H_0 \times W_0}$, we adopt the widely used ResNet \cite{he2016deep} to generate its low-resolution feature map $\mathbf{f}_{I} \in \mathbb{R}^{C_i \times HW}$. 
For the corresponding expression or sentence, we use the uncased base of BERT \cite{devlin2018bert} to obtain the representation $\mathbf{f}_{Q} \in \mathbb{R}^{C_t\times N}$, while $N$ is the length of the input context sentence. 

{\bf Visual-Lingual Encoder.}\quad
The visual-lingual encoder is designed to fuse information from multi-modal sources.
For cross-modality encoding, we use a transformer encoder based model, which is composed of 6 transformer encoder layers.
Specifically, given both image and text features, multi-layer perceptions are applied first to project different modalities to a joint embedding space with a hidden dimension of $C$.
Since transformer based encoders are permutation invariant and do not preserve positional information, we follow \citep{carion2020end, devlin2018bert} to add cosine positional embedding $P_{img}$ for image features and learnable positional embedding $P_{text}$ for text features.
We then concatenate the projected features into a single sequence $\mathbf{f} \in \mathbb{R}^{C\times (HW + N)}$. To distinguish between modalities, we also deign a learnable modal label embedding $E_{label}:\{E_{img}, E_{text}\}$ which is added to the original sequences.
The visual-lingual encoder then takes a sequence as input, and $\{P_{img}, P_{text}, E_{label}\}$ are fed into each encoder layer.
The encoder output is a multi-modal feature sequence $\mathbf{f}_{vl} \in \mathbb{R}^{C\times (HW + N)}$. 

\subsection{Referring Decoder}
\label{sec:approch_refer_decoder}
The referring decoder aims to decode the phrase query, conditioned on visual-lingual features from the encoder, into an output bounding box and segmentation. 
In this stage, we first generate query embeddings corresponding to each referring phrase. Then these query embeddings are fed into the decoder together with the joint embedding from the encoder to generate outputs.

{ \bf Encoding Referring Queries.}\quad
To enable the decoder to generate the desired output (bounding box and/or segmentation) the queries must encode several bits of crucial information. Mainly, (1) encoding of the referring phrase, (2) image encoding and (3) phrase-specific optional (con)text information. 
For phrase encoding in (1) we use a BERT model with pooling heads which share weights with (con)text encoder; this results in the phrase feature vector $\mathbf{f}_{\mathbf{p}_i} \in \mathbb{R}^C$ for the $i$-th referring phrase. 
We note that because we use visual-lingual encoder, (2) and (3) are jointly encoded in multi-modal features $\mathbf{f}_{vl}$ described in Section~\ref{sec:feature_extraction} above. 
However, $\mathbf{f}_{vl}$ is phrase-agnostic encoding of the image and (con)text. 
To generate phrase-specific context, given a phrase $\mathbf{p}_i$, average pooling is used to extract the phrase-specific context information $\mathbf{f}_{c}(\mathbf{p}_i)$ from the visual-lingual feature sequence as follows:
\begin{equation}
    \mathbf{f}_{c}(\mathbf{p}_i) = \frac{\sum \mathbf{f}_{vl}[l_{\mathbf{p}_i}:r_{\mathbf{p}_i}]}{r_{\mathbf{p}_i}-l_{\mathbf{p}_i}}
\end{equation}
where $l_{\mathbf{p}_i}$ and $r_{\mathbf{p}_i}$ denotes the left and right bounds of phrase $\mathbf{p}_i$ in the original (con)text sentence. Finally, given phrase encoding $\mathbf{f}_{\mathbf{p}_i}$ and phrase-specific context $\mathbf{f}_{c}(\mathbf{p}_i)$ we construct our phrase queries using a multi-later perceptron:
\begin{equation}
    \widehat{Q_{\mathbf{p}_i}} = {\tt MLP}\left([\mathbf{f}_{c}(\mathbf{p}_i); \mathbf{f}_{\mathbf{p}_i}]\right) + E_p, 
\end{equation}
where $E_p \in \mathbb{R}^C$ is a learnable embedding which serves as a bias to the formed query. 


{ \bf Decoding.}\quad
In the decoder, an attention graph convolution layer is used to enable information flow in a dense connected graph of phrase queries. This allows phrase queries to contextualize and refine each other; the inspiration for this step is taken from \cite{bajaj2019g3raphground}.
%
After that, a cross attention layer decodes visual-lingual information given the updated phrase query and feature sequence from the encoder.
The design of our decoder is similar to the transformer decoder, except attention is non-causal. 
%

\subsection{Multi-task training}
In this section, we demonstrate how the decoded phrase-specific query features can be naturally used to train multiple heads for different referring tasks (regression for REC and segmentation for RES). 

{ \bf Referring Comprehension/Detection (REC).}\quad
For referring detection tasks, the final output is computed by a simple two-layer perceptron over the decoded phrase-specific query features. We let the detection head directly output center coordinates $\tilde{\mathbf{b}} = (x, y, h, w)$ for the referred image. To supervise the training, we use a weighted sum of an L1 loss and a Generalized IOU loss \cite{rezatofighi2019generalized}: 
\begin{equation}
    \mathcal{L}_{det} = \lambda_{iou}\mathcal{L}_{iou}(\mathbf{b}, \tilde{\mathbf{b}}) + \lambda_{L1}||\mathbf{b}-\tilde{\mathbf{b}}||_1.
    \label{eq:loss_det}
\end{equation}
The $\lambda_{iou}$ and $\lambda_{L1}$ control the relative weighting of the two losses in the REC objective. 

{ \bf Referring Segmentation (RES).}\quad
Following previous work \cite{carion2020end}, we design an FPN-like architecture to predict a referring segmentation mask for each phrase expression. 
Attention masks from the decoder and image features from the visual-lingual encoder are concatenated as the FPN input, while features from different stages of image backbones are used as skip connections to refine the final output. 
The last linear layer project the upsampled feature to a single channel heatmap and a sigmoid function is used to map the feature to mask scores $\tilde{\mathbf{s}} \in \mathbb{R}^{H_0/4\times W_0/4}$. The loss for training RES task is:
\begin{equation}
    \mathcal{L}_{seg} = \lambda_{focal}\mathcal{L}_{focal}(\mathbf{s}, \tilde{\mathbf{s}}) + \lambda_{dice}\mathcal{L}_{dice}(\mathbf{s}, \tilde{\mathbf{s}}).
    \label{eq:loss_seg}
\end{equation}
Here $\mathcal{L}_{focal}$ is the focal loss for classifying pixels used in \cite{lin2017focal}, $\mathcal{L}_{dice}$ is the DICE/F-1 loss proposed in \cite{milletari2016v}; 
$\lambda_{focal}$ and $\lambda_{dice}$ are hyper-parameters controlling the relative importance of the two losses. 

{ \bf Joint Training.}\quad
While it is possible to train referring segmentation and referring detection tasks separately, we find that joint training is highly beneficial. Therefore the combined training loss which we optimize is $\mathcal{L} = \mathcal{L}_{seg} + \mathcal{L}_{det}$.

{ \bf Pretraining the Transformer.}\quad
\label{para:pretrain}
Transformers are generally data hungry and requires a lot of data to train \cite{chen2020uniter,lu2019vilbert}. Although in this paper we do not use a large pretraining model and a lot of data. We found that simple preatraining strategy on the region description splits of Visual-Genome dataset \cite{krishna2017visual} makes our model achieve comparable and even better performance against some of state-of-the-art pretrained models. 
Interestingly, we found that although there is no ground truth segmentation provided in Visual-Genome, the RES task can still benefit greatly from pretrained models, likely due to the fine-tuned multi-task representation.



\vspace{-0.07in}
\section{Experiments}
\vspace{-0.07in}
\label{sec:exp}
\subsection{Datasets}
\label{sec:dataset}
{\bf RefCOCO/RefCOCO+/RefCOCOg (REC\&RES).}\quad
RefCOCO, RefCOCO+ \cite{yu2016modeling} and RefCOCOg \cite{nagaraja2016modeling} are collections of images and referred objects from MSCOCO \cite{lin2014microsoft}. On RefCOCO and RefCOCO+ we follow the split used in \cite{yu2016modeling} and report scores on the validation, testA and testB splits. On RefCOCOg, we use the RefCOCO-umd splits proposed in\cite{nagaraja2016modeling}.

{\bf Flickr30k Entities (REC).}\quad
Flickr30k Entities \cite{plummer2015flickr30k} contains 31,783 images and 158k caption sentences with 427k annotated phrase.  We use splits from \cite{plummer2015flickr30k, plummer2018conditional}. Bounding boxes and phrase annotations are consistent with the previous one-stage approaches \cite{yang2019fast, yang2020improving} for fair comparisons.

{\bf ReferIt (REC).}\quad
The ReferItGame dataset \cite{kazemzadeh2014referitgame} contains 20,000 images. We follow setup in \cite{chen2017query} for splitting train, validation and test set; resulting in 54k, 6k and 6k referring expressions respectively.

\subsection{Implementing Details}
\label{sec:implementation}
We train our model with AdamW \cite{loshchilov2018decoupled}. The initial learning rate is set to 1e-4 while the learning rate of image backbone and context encoder is set to 1e-5. We initialized weights in the transformer encoder and decoder with Xavier initialization \cite{xavier_init_2010}. For image backbone, we experiment with the popular ResNet-50 and ResNet-101 networks \cite{he2016deep} where weights are initialized from corresponding ImageNet-pretrained models. For the context encoder and phrase encoder, we use an uncased version of BERT model \cite{devlin2018bert} with weights initialized from pretrained checkpoints provided by HuggingFace \cite{huggingface_transformers}. 
For data augmentation, we scale images such that the longest side is $640$ pixels and follow \cite{yang2019fast} to do random intensity saturation and affine transforms. We remove the random horizontal flip augmentation used in previous work \cite{yang2019fast} since we notice it causes semantic ambiguity on RefCOCO, likely due to relative location ({\em e.g.}, left of/right of) specific queries in the dataset.

On Flickr30k dataset, we set the maxium length of context sentence to 90 and maximum number of referring phrases to 16.
On the ReferIt and the RefCOCO dataset, only phrase expressions are provided and the task aims to predict a single bounding box for each of the expressions. In those cases, the context sentence is taken as the referring phrase expression itself. We set the maximum length of context sentence on these two datasets to 40.
To fairly compare with pretrained methods, we use region description split in the VisualGenome \cite{krishna2017visual} to pretrain our model. The dataset contains approximately 100k images and we remove the images that appears in Flickr30k Entities and RefCOCO/RefCOCOg/RefCOCO+'s validation and test set to avoid potential test data leak. For all the pretrained methods, we only train the model on pretraining dataset for 6 epoches. We find that longer pretraining schedule gives better performance, but since the focus of this paper is not on pretraining methods, we stick to shorter pretraining schedules to save computational resources.
All experiments are conducted using 4 Nvidia 2080TI GPU with batch size as 8. For all the results given, we run experiments several times with random seeds and the error bars are within $\pm0.5\%$. 
%


\subsection{Quantitative Analysis}
{\bf Evaluation Metrics.} \quad
For referring expression comprehension (REC), consistent with prior works, we use precision as the evaluation metric. We mark a referring detection as correct when the intersection-over-union (IoU) between the predicted bounding box and ground truth is larger than 0.5. For referring expression segmentation (RES), we reported the Mean IoU (MIoU) between the predicted segmentation mask and ground truth mask.

\begin{table}[htbp]
	\caption{{\bf Comparison on REC task.} Performance on  RefCOCO/RefCOCO+/RefCOCOg datasets \cite{yu2016modeling} is reported. Ours$^*$ denotes that pretraining is used. RN50 and RN101 refer to ResNet50 and ResNet101 \cite{he2016deep} respectively; DN53 refers to DarkNet53 \cite{redmon2018yolov3} backbone.}
	\vspace{-0.05cm}
	\small
	\begin{center}
	    \resizebox{0.95\textwidth}{!}{%
		\begin{tabular}{c| c|c|c| c c c | c c c | c c}
			\hline
			\multirow{2}{*}{Models}  & Visual & Pretrain & Multi- & \multicolumn{3}{c|}{RefCOCO} & \multicolumn{3}{c|}{RefCOCO+} & \multicolumn{2}{c}{RefCOCOg} \\ 
			& Features & Images & task & val & testA & testB & val & testA & testB & val-u & test-u \\
			\hline 
			\textit{Two-stage:} & & & & & & & & & & \\
			CMN~\cite{hu2017modeling}  & VGG16 & None & $\xmark$ & - & 71.03 & 65.77 & - & 54.32 & 47.76 & - & - \\
			MAttNet~\cite{yu2018mattnet} &	RN101 & None & $\xmark$ & 76.65 & 81.14 & 69.99 & 65.33 & 71.62 & 56.02 & 66.58 & 67.27 \\
			RvG-Tree~\cite{hong2019learning}  & RN101 & None & $\xmark$ & 75.06 & 78.61 & 69.85 & 63.51 & 67.45 & 56.66 & 66.95 & 66.51 \\
			NMTree~\cite{liu2019learning} & RN101 & None & $\xmark$ & 76.41 & 81.21 & 70.09 & 66.46 & 72.02 & 57.52 & 65.87 & 66.44 \\
			CM-Att-Erase~\cite{liu2019improving} &	RN101 & None & $\xmark$ & 78.35 & 83.14 & 71.32 & 68.09 & 73.65 & 58.03 & 67.99 & 68.67\\
			\hline
			\textit{One-stage:} & & & & & & & & & & &\\
			RCCF~\cite{liao2020real} & DLA34 & None & $\xmark$ & - & 81.06 & 71.85 & - & 70.35 & 56.32 & - & 65.73 \\
			SSG~\cite{chen2018real}  & DN53 & None & $\xmark$ & - & 76.51 & 67.50 & - & 62.14 & 49.27 & 58.80 & - \\  
			FAOA~\cite{yang2019fast} & DN53 & None & $\xmark$ & 72.54 & 74.35 & 68.50 & 56.81 & 60.23 & 49.60 & 61.33 & 60.36 \\
			ReSC-Large~\cite{yang2020improving} & DN53 & None & $\xmark$ & 77.63 & 80.45 & 72.30 & 63.59 & 68.36 & 56.81 & 67.30 & 67.20 \\
			MCN ~\cite{luo2020multi}& DN53 & None &$\cmark$ & 80.08 & 82.29 & 74.98 & 67.16 & 72.86 & 57.31 & 66.46 & 66.01 \\
            \hline
			Ours & RN50 & None & $\cmark$ & \underline{81.82} & \underline{85.33} & \underline{76.31} & \underline{71.13} & \underline{75.58} & \underline{61.91} & \underline{69.32} & \underline{69.10} \\ 
			Ours & RN101 & None & $\cmark$ & \textbf{82.23} & \textbf{85.59} & \textbf{76.57} & \textbf{71.58} & \textbf{75.96} & \textbf{62.16} & \textbf{69.41} & \textbf{69.40} \\ 
            \hline
            \hline
            \textit{Pretrained:} & & & & & & & & & & &\\
			VilBERT\cite{lu2019vilbert} & RN101 & 3.3M & $\xmark$ & - & - & - & 72.34 & 78.52 & 62.61 & - & - \\
			ERNIE-ViL\_L\cite{yu2020ernie} & RN101 & 4.3M & $\xmark$ & - & - & - & 75.89 & 82.37 & 66.91 & - & - \\
			UNTIER\_L\cite{chen2020uniter} & RN101 & 4.6M & $\xmark$ & 81.41 & 87.04 & 74.17 & 75.90 & 81.45 & 66.70 & 74.86 & 75.77 \\
			VILLA\_L\cite{gan2020large} & RN101 & 4.6M & $\xmark$ & 82.39 & \underline{87.48} & 74.84 & 76.17 & \underline{81.54} & 66.84 & 76.18 & 76.71 \\
			\hline
			Ours* & RN50 & 100k & $\cmark$ & \underline{85.43} & \underline{87.48} & \underline{79.86} & \underline{76.40} & 81.35 & \underline{66.59} & \underline{78.43} & \underline{77.86}  \\
			Ours* & RN101 & 100k & $\cmark$ & \textbf{85.65} & \textbf{88.73} & \textbf{81.16} & \textbf{77.55} & \textbf{82.26} & \textbf{68.99} & \textbf{79.25} & \textbf{80.01} \\
			\hline
		\end{tabular}
	    }
	\end{center}
	\label{tab:refcoco_results}
    \vspace{-0.1in}
\end{table}

\begin{table}[htbp]
\footnotesize
\caption{{\bf Comparison on RES tasks.} Performance on  RefCOCO/RefCOCO+/RefCOCOg datasets \cite{yu2016modeling} is reported.
Ours$^*$ denotes that pretraining is used. RN50 abd RN101 refer to ResNet50 and ResNet101 \cite{he2016deep} respectively; DN53 refers to DarkNet53 \cite{redmon2018yolov3} backbone.}
\small
\begin{center}
    \resizebox{0.9\textwidth}{!}{ 
        \begin{tabular}{l|c|c c c|c c c|c c| c}
            \hline
            \multirow{2}{*}{Methods}&\multirow{2}{*}{Backbone}&\multicolumn{3}{c|}{RefCOCO}&\multicolumn{3}{c|}{RefCOCO+} &\multicolumn{2}{c|}{RefCOCOg} & Inference \\
             &  & val & testA & testB & val & testA & testB & val & test & time(ms) \\
            \hline
            DMN \cite{margffoy2018dynamic} & RN101 & 49.78 & 54.83 & 45.13 & 38.88 & 44.22 & 32.29 & - & - & - \\
            MAttNet \cite{yu2018mattnet} & RN101 & 56.51 & 62.37 & 51.70 & 46.67 & 52.39 & 40.08 & 47.64 & 48.61 & 378 \\
            NMTree \cite{liu2019learning} & RN101 & 56.59 & 63.02 & 52.06 & 47.40 & 53.01 & 41.56 & 46.59 & 47.88 & - \\
            Lang2seg \cite{chen2019referring} & RN101 & 58.90 & 61.77 & 53.81 & - & - & - & 46.37 & 46.95 & - \\
            BCAM \cite{hu2020bi} & RN101 & 61.35 & 63.37 & 59.57 & 48.57 & 52.87 & 42.13 & - & - & - \\
            CMPC \cite{huang2020referring} & RN101 & 61.36 & 64.53 & 59.64 & 49.56 & 53.44 & 43.23 & - & - & - \\
            CGAN \cite{luo2020cascade} & DN53 & 64.86 & 68.04 & 62.07 & 51.03&55.51&44.06&51.01 & 51.69 & - \\
            LTS \cite{jing2021locate}& DN53 & 65.43 & 67.76 & 63.08 & 54.21 & 58.32 & 48.02 & 54.40 & 54.25 &-\\
            MCN+ASNLS \cite{luo2020multi} & DN53 & 62.44 & 64.20 & 59.71 & 50.62 & 54.99 & 44.69 & 49.22 & 49.40 & 56 \\
            \hline
            Ours  & RN50 & 69.94 & 72.80 & 66.13 & 60.9 & 65.20 & 53.45 & 57.69 & 58.37 & \textbf{38} \\
            Ours  & RN101 & 70.56 & 73.49 & 66.57 & 61.08 & 64.69 & 52.73 & 58.73 & 58.51 & \underline{41} \\
            Ours$^*$ & RN50 & \underline{73.61} & \underline{75.22} & \underline{69.80} & \underline{65.30} & \underline{69.69} & \underline{56.98} & \underline{65.70} & \underline{65.41} & \textbf{38} \\
            Ours$^*$ & RN101 & \textbf{74.34} & \textbf{76.77} & \textbf{70.87} & \textbf{66.75} & \textbf{70.58} & \textbf{59.40} & \textbf{66.63} & \textbf{67.39} & \underline{41} \\
            \hline
        \end{tabular}
    }
\end{center}
\label{table:refcoco_SEG_result}
\vspace{-0.1in}
\end{table}

{\bf REC and RES on Refcoco/Refcoco+/Refcocog.}\quad
Our model addresses REC and RES tasks jointly.
We compare their respective performances with 
the state-of-the-art in Table \ref{tab:refcoco_results} and Table \ref{table:refcoco_SEG_result}. In Table \ref{tab:refcoco_results}, the model is first compared with previous one-stage and two-stage approaches for REC. Without bells and whistles, we observe a consistent performance boost of $+2.7\%/+4\%/+2.1\%$ on RefCOCO, $+6.6\%/+4.3\%/+8.5\%$ on RefCOCO+ and $+4.4\%/+5.1\%$ on RefCOCOg. 
%
To compare with pretrained BERT methods, we use the pretraining strategy discussed in Section~\ref{para:pretrain}. As results show, our model achieves comprehensive advantage and shows distinct improvement on some splits, even compared to advance BERT models that use $40 \times$ more data in pretraining.
%
Table \ref{table:refcoco_SEG_result} illustrates results on RES task in terms of MIoU score. It can be seen that our model achieves the best performance; substantially better than the state-of-art.  
We further observe that pretraining on the REC task gives a huge performance boost to the RES task, even when no segmentation mask is used in pre-training. Multi-task training enables the model to leverage performance boost in one task to improve the other. 

We show the inference time for our model in Table~\ref{table:refcoco_SEG_result}. 
Our model can directly decode all referring queries in an image in parallel, allowing it to reach real-time performance. 
%
Importanly, note the corresponding scores for our model in Table \ref{tab:refcoco_results} and Table \ref{table:refcoco_SEG_result} are based on the output of a single multi-task model that predicts referring detection box and segmentation mask simultaneously. The only related work that shares this property is the MCN \cite{luo2020multi}, which has substantially inferior performance. 
%
\label{sec:quali_result}
\begin{figure}[t]
  \centering
  \includegraphics[width=0.95\textwidth]{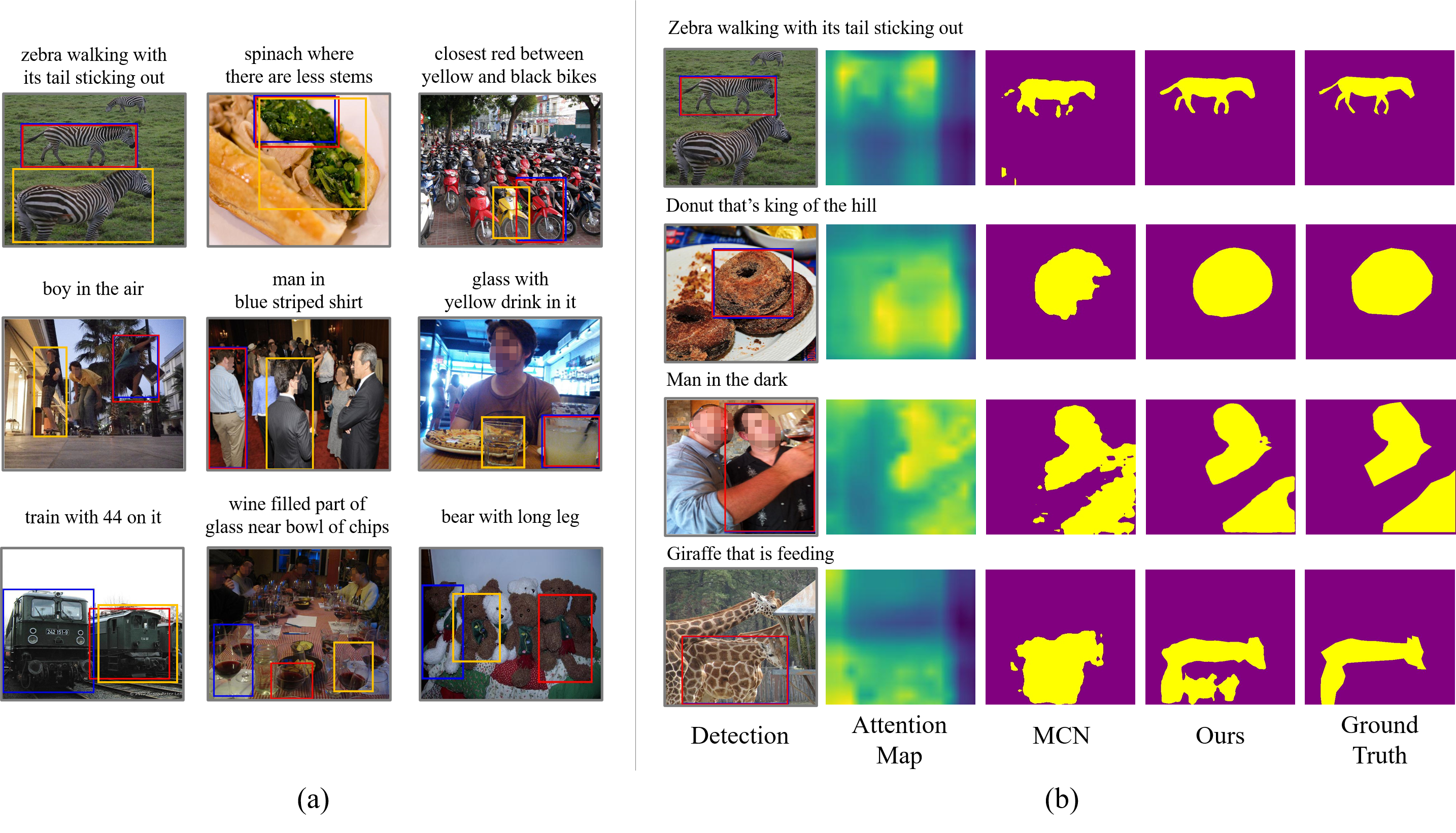}
  \caption{{\bf Qualitative Evaluation.} 
  In (a) comparison to MCN \cite{luo2020multi} on REC is shown; orange, blue and red bounding boxes correspond to outputs from MCN, our model and the ground truth. In (b) similar comparison on RES is made. 
  The attention map is drawn from the last layer of the decoder. We add mosaic to all human face to protect personal information.}
  \label{fig:quali_result_refcoco} 
\vspace{-0.05in}
\end{figure}

\begin{table}[htbp]
	\caption{{\bf Comparison with State-of-The-Art Methods.} Table illustrates performance on the test set of  ReferItGame~\cite{kazemzadeh2014referitgame} and Flickr30K Entities~\cite{plummer2015flickr30k} datasets in terms of top-1 accuracy (\%).}
	\small
	\begin{center}
		\resizebox{0.6\textwidth}{!}{
			\begin{tabular}{c | c | c | c | c }
				\hline
				\multirow{2}{*}{Models} & \multirow{2}{*}{Backbone} & \multicolumn{1}{c|}{ReferItGame} & Flickr30K & Inference time\\
				& & test & test & on Flickr30k(ms)\\
				\hline
				\multicolumn{5}{c}{{\textit{Two-stage}}}\\
				\hline
				MAttNet~\cite{yu2018mattnet} &	RN101  & 29.04 & - & 320\\
				Similarity Net~\cite{wang2018learning} & RN101 & 34.54 & 60.89 & 184\\
				CITE~\cite{plummer2018conditional}  & RN101  & 35.07 & 61.33 & 196\\
				DDPN~\cite{yu2018rethinking} & RN101 &  63.00 & 73.30 & - \\
				\hline
				\hline
				\multicolumn{5}{c}{{\textit{One-stage}}}\\
				\hline
				SSG~\cite{chen2018real} & DN53 & 54.24 & - & 25\\
				ZSGNet~\cite{sadhu2019zero} & RN50 & 58.63 & 58.63 & -\\
				FAOA~\cite{yang2019fast} & DN53 & 60.67 & 68.71 & 23\\
				RCCF~\cite{liao2020real} & DLA34 & 63.79 & - & 25 \\
				ReSC-Large~\cite{yang2020improving} & DN53 & 64.60 & 69.28 & 36\\
				\hline
				Ours & RN50 & 70.81 & 78.13 & 37(14)\\
				Ours & RN101 & 71.42 & 78.66 & 40(15)\\
				Ours* & RN50 & \underline{75.49} & \underline{79.46} & 37(14)\\
				Ours* & RN101 & \textbf{76.18} &  \textbf{81.18} & 40(15)\\
				\hline
			\end{tabular}
		} 
	\end{center}
	\label{tab:flickr_result}
\vspace{-0.2in}
\end{table}

{\bf REC on Flickr30k-Entities.}\quad
For Flickr30k dataset, previous one-stage works \cite{yang2019fast, yang2020improving} extract short phrases from sentences and treated them as separate referring expression comprehension tasks. We argue that in this setting, queries are mostly short phrases and therefore cannot well reflect the model's ability to comprehend them in context. In contrast, our model, given an image and a caption (con)text sentence, aims to predict bounding boxes for {\em all} referred entities in the sentence. 
%
Doing so gives several advantages: 1. We are able to contextualize referring expressions given all other referring expressions and (con)text provided by the sentence. 2. Locations for all phrases can be inferred in one forward pass of the network, which saves a lot of computation as compared to previous one-stage approaches \cite{yang2019fast,yang2020improving} that process one phrase at a time. Note that our task formulation is consistent with some two-stage models \cite{bajaj2019g3raphground,dogan2019neural}, but is unique for a one stage approach. 

In Table~\ref{tab:flickr_result} we compare our results with state-of-the-art methods. Without pretraining, we obtain a huge performance boost compared to both previous one-stage (+13.54\%) and two-stage (+7.31\%) state-of-the art methods. By using pretrained models, we observe that our model tends to generalize better on the test set and gives even better performance. 
We also provide comparison of inference time both per image and (per-expression), since our model can amortize inference across expressions. Per-expression, our inference time is substantially lower than all prior methods.

{\bf REC on ReferIt.}\quad
Since ReferIt is a relative small dataset, we use a slightly smaller model which contains 3 cross attention layers in the query decoder. The results are shown in Table~\ref{tab:flickr_result}. Our model is able to perform better, by a large margin, than even the latest one-stage methods. We also observe a consistent boost brought by pretraining.

\subsection{Qualitative Analysis}

In Figure \ref{fig:quali_result_refcoco}, we show our qualitative comparison with previous state-of-the-art multi-task model -- MCN \cite{luo2020multi}. The first two rows of Figure~\ref{fig:quali_result_refcoco}(a) show failure cases of MCN that can be better handled by our model. We observe that MCN appears to fail because it neglects some attributes in referring expression ({\em e.g.}, "\textit{yellow} drink" and "\textit{blue} striped shirt"), while our model is able to better model the query and pay attention to object attributes.
In the last row, we shows several failure cases of our model. For the first case, the query requires the model to have the ability to recognize number "44". For the second and third case, there is visual ambiguity to identify the nearest glass to the bowl or to determine which bear(brown or white) has the longest leg.

In Figure~\ref{fig:quali_result_refcoco}(b), we show qualitative comparison in terms of referred segmentation mask. Compared to MCN, our model is able to output more detailed object shape and finer outlines. Moreover, our model shows the ability to handle shadows ({\em e.g.}, the right bottom of the donut) and occlusions ({\em e.g.}, the man occluded by another man's arm) and predict smoother segmentation mask.
We also give a result on a challenging case in the last row, where the texture boundary of the two giraffe is hard to distinguish. Despite imperfections, our model is still able to focus on the giraffe's head in the foreground and performs much better than MCN. 


\subsection{Ablation Studies}
\label{sec:ablation}
\begin{table}[t]\small
\centering
\caption{{\bf Ablation studies.} Table on the left ablates our multi-task and pretraining schehme on RefCOCO+ validation set. Table on the right ablates on core components of our model.}
\label{tab:ablation}
\parbox{.45\linewidth}{
\centering
\resizebox{.45\textwidth}{!}{
    \begin{tabular}{ |c c c|c c c| }
        \hline
         REC & RES & \multirow{2}{*}{Pretrain} & \multicolumn{3}{c|}{RefCOCO+}\\
         Loss & Loss &  & REC$\uparrow$ & RES$\uparrow$ & IE$\downarrow$\\
        \hline
                & $\cmark$ & & - & 58.39 & \multirow{2}{*}{23.52\%}\\
         $\cmark$ &        & & 70.02 & - & \\
         $\cmark$ & $\cmark$ & & 71.13 & 61.08 & 4.73\%\\
         \hline
         $\cmark$ &        & $\cmark$ & 75.07 &  - & \\
         $\cmark$ & $\cmark$ & $\cmark$ & 76.40 & 65.30 & 4.48\%\\
        \hline
        \end{tabular}
    }
}
\quad
\quad
\parbox{.45\linewidth}{
\centering
\resizebox{.4\textwidth}{!}{
    \begin{tabular}{|l| c | c|}
        \hline
         & Flickr30k \\
        \hline
         \multicolumn{1}{|c|}{Model component} & \\
         -w/o \textit{Query Decoder}  & 49.38 \\
         -w/o \textit{Context Encoder} & 73.68 \\
         \hline
         \multicolumn{1}{|c|}{Query Encoder}  & \\
         -w/o \textit{Context \& Phrase Feature} & 42.05\\
         -w/o \textit{Context Feature}  & 76.64 \\
         -w/o \textit{Phrase Feature}  & 77.02 \\
         -w/o \textit{Learnable Embedding}  & 77.64 \\
         \hline
         \multicolumn{1}{|c|}{\textit{Full model}} & 78.13 \\
        \hline
    \end{tabular}
}
}
\vspace{-0.15in}
\end{table}

We first consider the importance of the multi-task setup in Table \ref{tab:ablation} (left). 
Results indicate that multi-task training boosts both REC and RES performance by a considerable margin. More specifically, we observe that REC loss helps the transformer to better locate the referred object and converge faster in early stages of training. At the same time, RES loss aids the model with more fine-grained information on the shape of the referred region, which helps to further enhance the accuracy.
IE here is \textit{Inconsistency Error} metric originally used in \cite{luo2020multi} to measure the prediction conflict between the REC and RES task.
We can see that joint training of RES and REC greatly reduce the inconsistency between the two tasks. 
Note that our model also has a much lower multi-task inconsistency compared to MCN \cite{luo2020multi}, with a corresponding IE score of 7.54\%(-40\%).
This shows that our model can do better collaborative learning. 

Next, we validate the design of our network architecture. We report our scores on the Flickr30k test sets. In Table~\ref{tab:ablation} (right), we ablate the model's major components and features used to form the query. Without (w/o) context encoder indicates that we directly use learnable embedding to encode text; w/o Query Decoder means that we directly use the average pooled feature from the encoder to predict a single referred output. We can see that the context encoder plays an important role in providing good textual representation for further multi-modal fusion. Query encoders are also quite important without which we also observe a big performance drop.
For the ablation on query features, we observe that both context feature and phrase features are crucial without which the performance will decrease considerably. 
Learnable embeddings also help to enhance the performance by adding bias to fused query semantics. 
The table also showed that the network will not work without guidance of both context and phrase features since we cannot establish a correspondence between multiple queries and outputs in such a case.


\vspace{-0.07in}
\section{Conclusions and Future Work}
\vspace{-0.07in}
\label{sec:conclusion}
In this work, we present Referring Transformer, a one-step approach to referring expression comprehension (REC) and segmentation (RES). We jointly train our model for RES and REC tasks while enabling contextualized multi-expression references. Our models outperform state-of-the-art by a large margin on five / three datasets for REC / RES respectively, while achieving real-time runtime. 

One limitation for our model is that we follow the setup in previous works~\cite{yang2019fast, yang2020improving} and assume that each expression refers to only one region. In the future, we plan to explore 
learning to predict multiple regions for each referring entity if necessary. 
Large-scale multi-task pretraining has been demonstrated to be very effective for ViLBEERT and other similar architectures; this is complementary to our focus in this paper, and we expect such strategies to further improve the performance.

\section{Acknowledgments and Disclosure of Funding}

This work was funded, in part, by the Vector Institute for AI, Canada CIFAR AI Chair, NSERC CRC and an NSERC DG and Discovery Accelerator Grants.
Resources used in preparing this research were provided, in part, by the Province of Ontario, the Government of Canada through CIFAR, and companies sponsoring the Vector Institute \url{www.vectorinstitute.ai/#partners}.
Additional hardware support was provided by John R. Evans Leaders Fund CFI grant and Compute Canada under the Resource Allocation Competition award.

\newpage 
{\small
\bibliographystyle{abbrvnat}
\bibliography{egbib}
}

\newpage 
\appendix
\section{More Details for Referring Expression Segmentation (RES)}
\begin{figure}[htbp]
  \centering
  \includegraphics[width=0.98\textwidth]{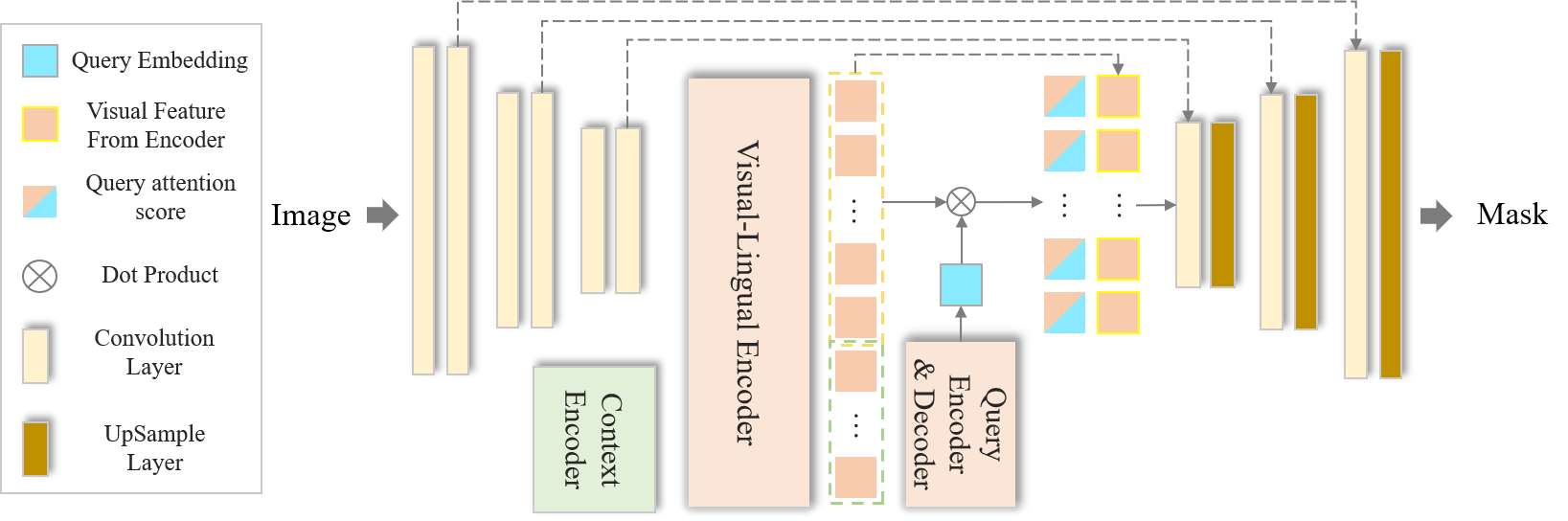}
  \caption{{\bf RES Task Head.} A detailed illustration of our model for RES task.}
  \label{fig:net_RES} 
\end{figure}
We provide a more detailed illustration of our model for the RES task in Figure~\ref{fig:net_RES}.
With decoded query embedding from the query decoder and visual feature coming from visual-lingual decoder, a query attention score $\mathbf{S}_{att} \in \mathbb{R}^{M\times (HW)}$ is computed using a dot product. Here $M$ denotes the number of attention heads, which is $8$ in our implementation. The attention score is then concatenated with visual feature and sent into several up-sampling blocks (convolution layer with stride of 2). We also add residual connections from different stages of the ResNet backbone to help refine the up-sampled features. All convolution layers here use a kernel size of 3. The design is motivated by MaskRCNN\cite{mask_rcnn} and DETR\cite{carion2020end}.

\section{Additional Implementation Details}

{\bf Pretraining.} \quad
We use the description split of Visual Genome~\cite{krishna2017visual} for pretraining, it contains 100k images with an average of 40 region descriptions per-image. 
We pretrain our model with REC task on the Visual Genome dataset for 6 epochs. We set the learning rate at 1e-4 and decay it by 10x after 4 epochs. The trained model is then used to initialize the model for dataset-specific fine-tuning.

{\bf RefCOCO Training.} \quad
For experiments on RefCOCO(+/g) \cite{nagaraja2016modeling,yu2016modeling}, since auxiliary loss is expensive for RES task, we first train our model with auxiliary loss on REC task for 60 epochs using a learning rate of 1e-4. Then, we disable auxiliary loss and train the model jointly on RES and REC task for 30 epochs with learning rate of 1e-4 that decays on the 10th epoch.

{\bf ReferItGame / Flickr30k Training.} \quad
On ReferItGame \cite{kazemzadeh2014referitgame} and Flickr30k Entities \cite{plummer2015flickr30k}, our model is trained for 90 and 60 epochs respectively, with learning rate decays on the 60th and 40th epoch.

{\bf Source Code.} \quad
We include core codes for our model in the Supplemental for the reference. We will release complete code with checkpoints to reproduce reported scores upon acceptance.

\section{Additional Results}

\begin{figure}[t!]
  \centering
  \includegraphics[width=0.98\textwidth]{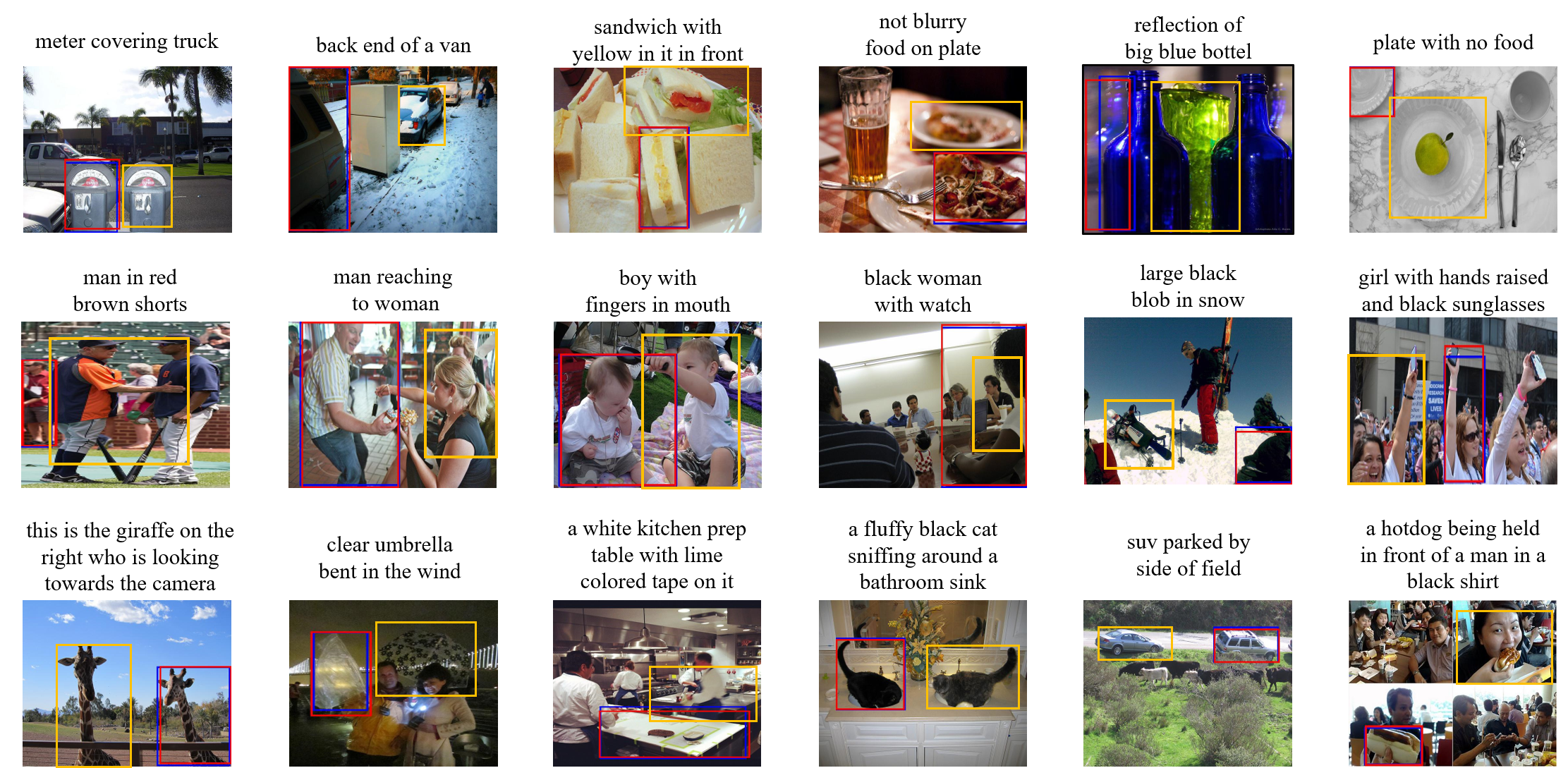}
  \caption{{\bf Additional Qualitative Results on REC Task.} Orange, blue and red bounding boxes correspond to outputs from MCN~\cite{luo2020multi}, our model and the ground truth. The first row, second row and third row comes from RefCOCO+ testA, testB and RefCOCOg test set respectively.}
  \label{fig:qualitative_rec_supp} 
\end{figure}

\begin{figure}[t!]
  \centering
  \includegraphics[width=0.98\textwidth]{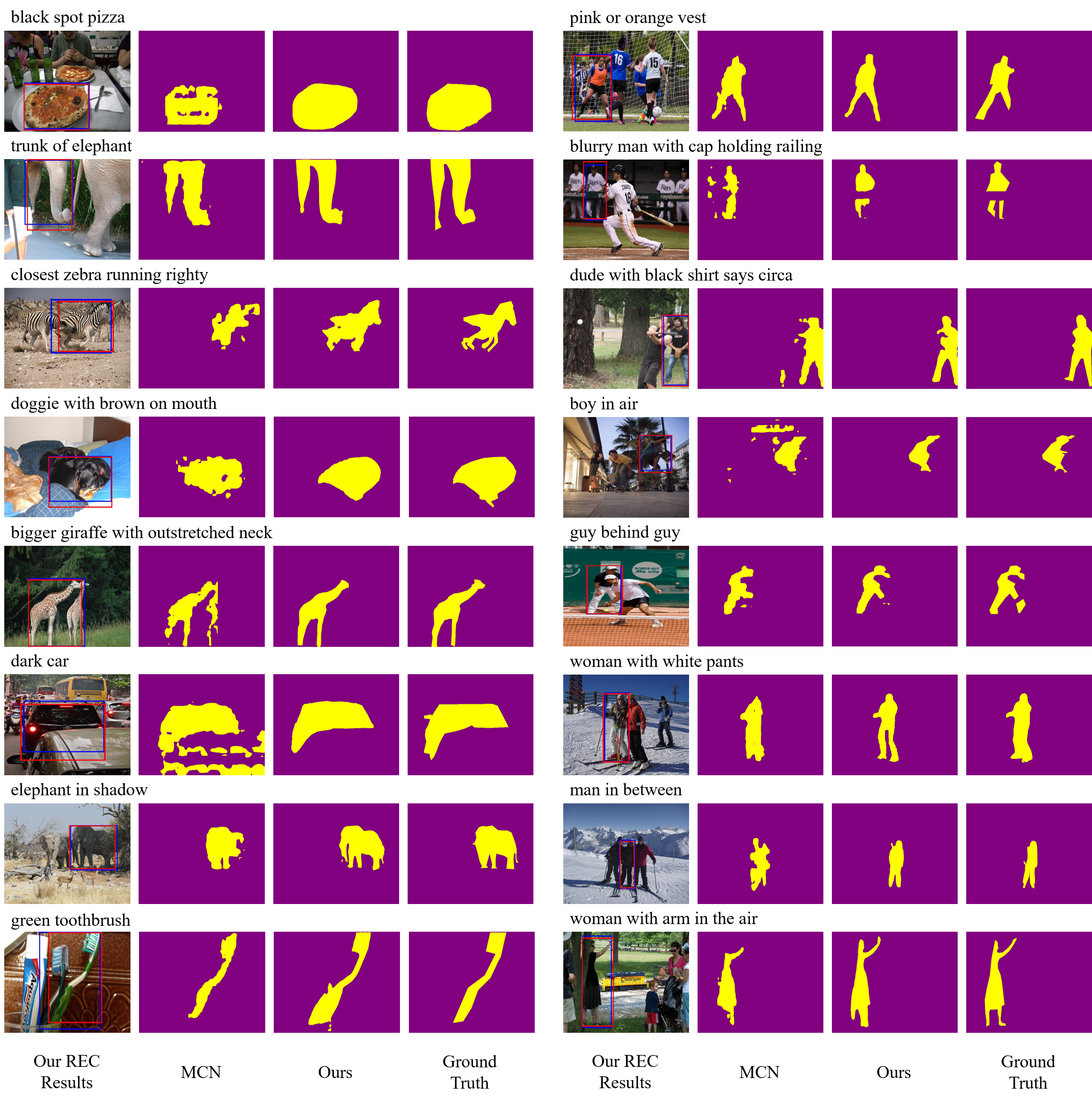}
  \caption{{\bf Additional Results on RES Task.} Images come from RefCOCO+ testA and testB splits.}
  \label{fig:qualitative_res_supp} 
\end{figure}

{\bf More Qualitative Comparison.} \quad
Additional qualitative results on REC and RES tasks, compared to the previous multi-task framework of \cite{luo2020multi}, are shown in Figure~\ref{fig:qualitative_rec_supp} and Figure~\ref{fig:qualitative_res_supp} respectively. Note that the score of RES can benefit greatly from better localization of corresponding REC task. In Figure~\ref{fig:qualitative_res_supp}, to better compare the quality of generated referred masks, we compare the mask quality in the case where both MCN~\cite{luo2020multi} and our method assume correct REC localization.

{\bf Results with Different Input Resolution.} \quad
In RES and REC tasks, the size of input image is a matter of trade off between performance and speed, which is largely effected by the network architecture. Despite that our model is designed to be able to process $640 \times 640$ images at real time speed, we provide our model with different input resolution for reference, as showed in Table~\ref{tab:resolution}. Note that we replace convolution in the final stage with dilated convolution to make sure the number of visual features sent into visual-lingual encoder are roughly the same.
\begin{table}[t]\small
\centering
\caption{{\bf Results on RefCOCO+ Dataset with Different Input Resolutions.} Our methods correspond to the model without pretraining.}
\vspace{-6pt}
\centering
\resizebox{.85\textwidth}{!}{
    \begin{tabular}{|c|c|c c c|c c c|}
        \hline
        Models & Resolution & \multicolumn{3}{c|}{REC(prec@0.5)} & \multicolumn{3}{c|}{RES(MIoU)}\\
        & & val & testA & testB & val & testA & testB\\
        \hline
        FAOA~\cite{yang2019fast} & 256$\times$ 256 & 56.81 & 60.23 & 49.60 & - & - & - \\
        ReSC-Large~\cite{yang2020improving} & 256$\times$ 256& 63.59 & 68.36 & 56.81 & - & - & - \\
        \hline
        CMPC~\cite{huang2020referring} & 320$\times$ 320 & - & - & - & 49.56 & 53.44 & 43.23 \\
        LTS~\cite{jing2021locate} & 416$\times$ 416& - & - & - & 54.21 & 58.32 & 48.02 \\
        MCN~\cite{luo2020multi} & 416$\times$ 416 & 67.16 & 72.86 & 57.31 & 50.62 & 54.99 & 44.69\\
        \hline
        Ours & 256$\times$ 256& 70.05 & 73.29 & 61.48 & 58.26 & 61.09 & 52.20\\
        Ours & 320$\times$ 320& 70.03 & 73.23 & 61.52 & 58.42 & 61.48 & 52.34\\
        Ours & 416$\times$ 416& 71.50 & 75.87 & 61.71 & 61.00 & 64.48 & 52.44\\
        Ours & 640$\times$ 640& 71.58 & 75.96 & 62.16 & 61.08 & 64.69 & 52.73\\
        \hline
        \end{tabular}
    }
\vspace{-10pt}
\label{tab:resolution}
\end{table}

{\bf Comparison with Contemporaneous Work}\quad
As discussed in the main paper,~\cite{deng2021transvg, du2021visual, kamath2021mdetr} are  non-refereed/unpublished works that appear on ArXiv recently.
We provide quantitative comparison on REC task with these approaches, based on their reported numbers in Table~\ref{tab:refcoco_results}. We stress that {\em none} of these works address or show performance on multi-task performance of REC \& RES task, which is one of distinctive qualities of our model in comparison. 

Compared to \cite{deng2021transvg, du2021visual}, our model performs substantially better in all, with an exception of RefCOCO testB (where \cite{deng2021transvg} is marginally better), datasets and splits. 
The biggest improvements can be seen on RefCOCO+, where our model is 10.4\% better (or 6.76 points better), than the closest concurrent work of \cite{deng2021transvg}, on the Val split; similar sizable improvements are illustrated on other splits, {\em e.g.}, 9.2\% on RefCOCO+ testB. 
In addition, our approach is considerably faster in runtime, since our model is able to handle multiple queries simultaneously (unlike \cite{deng2021transvg, du2021visual}). Unfortunately, we are not able to report specific numbers for this at this time, as \cite{deng2021transvg, du2021visual} do not report them in their ArXiv papers.

Compared to \cite{kamath2021mdetr}, in a pretrained model setting, we see that our model performs similarly to \cite{kamath2021mdetr} on RefCOCO and marginally worse on RefCOCO+ and RefCOCOg. We believe this difference can be attributed to two factors: (1) pretraining on a larger dataset  (200k images vs. 100k for us, plus we use a shorter 6 epochs pretraining schedule) (2) using more sophisticated language model (RoBERTa for \cite{kamath2021mdetr} vs. vanilla BERT for us). We were unable to explore effect those choices would have on our model for the moment, but expect them to further boost the performance. In addition, we setup our method in multi-task setting to solve RES and REC task at the same time, so our formulation while perhaps marginally inferior on REC is more general overall. Finally, our formulation, which does away with Hungarian matching loss, is likely to also be significantly faster to train. Exploring this would require re-running \cite{kamath2021mdetr}, which we hope to do for the camera ready.


\begin{table}[htbp]
	\caption{{\bf Comparison with Concurrent Work.} Performance on  RefCOCO/RefCOCO+/RefCOCOg datasets \cite{yu2016modeling} is reported. Ours$^*$ denotes that pretraining is used. All methods use ResNet101 as the image backbone.}
	\vspace{-0.05cm}
	\small
	\begin{center}
	    \resizebox{0.95\textwidth}{!}{ 
		\begin{tabular}{c| c|c|c| c c c | c c c | c c}
			\hline
			\multirow{2}{*}{Models}  & Visual & Pretrain & {Multi-} & \multicolumn{3}{c|}{RefCOCO} & \multicolumn{3}{c|}{RefCOCO+} & \multicolumn{2}{c}{RefCOCOg} \\ 
			& Features & Images & task & val & testA & testB & val & testA & testB & val-u & test-u \\
			\hline 
			\textit{One-stage:} & & & & & & & & & & &\\
			VGTR~\cite{du2021visual} & Bi-LSTM & None & $\xmark$ & 79.20 & 82.32 & 73.78 & 63.91 & 70.09 & 56.51 & 65.73 & 67.23\\
			TransVG~\cite{deng2021transvg} & BERT & None & $\xmark$ & 81.02 & 82.72 & {78.35} & 64.82 & 70.70 & 56.94 & 68.67 & 67.73\\
            \hline
			Ours & BERT & None & $\cmark$ & {82.23} & {85.59} & 76.57 & {71.58} & { 75.96} & {62.16} & {69.41} & {69.40} \\ 
            \hline \hline
            \textit{Pretrained:} & & & & & & & & & & &\\
			MDETR~\cite{kamath2021mdetr} & RoBERTa & 200k & $\cmark$ & 86.75 & 89.58 & 81.41 & 79.52 & 84.09 & 70.62 & 81.64 & 80.89\\
			\hline
			Ours$^*$ & BERT & 100k & $\cmark$ & 85.65 & 88.73 & 81.16 & 77.55 & 82.26 & 68.99 & 79.25 & 80.01 \\
			\hline
		\end{tabular}
	}
	\end{center}
	\label{tab:refcoco_results}
    \vspace{-0.1in}
\end{table}


\end{document}